\documentclass[preprint]{acm_proc_article-sp}
\usepackage{epsfig}
\usepackage[square,comma,numbers,sort&compress]{natbib}
\usepackage{multirow}
\newcommand{\UPLB}{University of the Philippines Los Ba\~{n}os}
\begin{document}

\title{Neural Network Classifiers for Natural Food Products\titlenote{This paper is an improved paper that was judged {\it Finalist} in the {\bf 2008 PCASTRD Outstanding Research and Development in Advanced Science and Technology}, Heritage Hotel, Pasay City, 18 December 2008. The paper was only seen by the 5 members of the board of judges and has not been presented publicly.}}
\numberofauthors{1}
\author{
\alignauthor Jaderick P. Pabico, Alona V. De Grano and Alan L. Zarsuela\\
   \affaddr{Institute of Computer Science}\\
   \affaddr{\UPLB}\\
   \email{jppabico@uplb.edu.ph}
}
\date{}
\toappearbox{Contributed scientific paper to the 2012 Philippine Computing Science Congress, De La Salle Canlubang, Bi\~nan, Laguna, 1--3 March 2012.}

\maketitle

\begin{abstract}
Two cheap, off-the-shelf machine vision systems (MVS), each using an artificial neural network (ANN) as classifier, were developed, improved and evaluated to automate the classification of tomato ripeness and   acceptability of eggs, respectively. Six thousand color images of human-graded tomatoes and 750 images of human-graded eggs were used to train, test, and validate several multi-layered ANNs. The ANNs output the corresponding grade of the produce by accepting as inputs the spectral patterns of the background-less image. In both MVS, the ANN with the highest validation rate was automatically chosen by a heuristic and its performance compared to that of the human graders'. Using the validation set, the MVS correctly graded 97.00\% and 86.00\% of the tomato and egg data, respectively. The human grader's, however, were measured to perform at a daily average of 92.65\% and 72.67\% for tomato and egg grading, respectively. This results show that an ANN-based MVS is a potential alternative to manual grading.\end{abstract}


\section{Introduction}
\begin{figure}[hbt]
\centering\epsfig{file=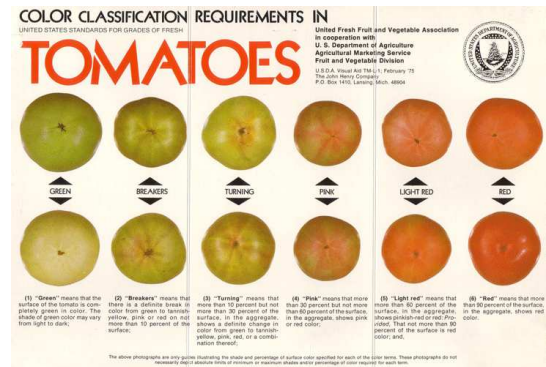, width=2.8in}
\caption{The USDA Color Chart for the classification of ripeness in tomatoes~\citep{USDAColorChart}.}\label{fig:USDA}
\end{figure}

The manual procedure of sorting or classifying objects through the objects' various visual characteristics is a very difficult process to model, specifically if the procedure was learned through experience and trial-and-error. Because of this difficulty, automating the manual process for the purpose of providing efficiency and preservation of the human expertise has also become a very difficult problem to solve. In the Philippines, the most common process of classifying tomato ({\it Lycopersicon esculentum}) maturity is through visual comparison of the fruit's skin color with the United States Department of Agriculture (USDA) Color Chart (Figure~\ref{fig:USDA}). Color in tomatoes is the most important external characteristics to assess its ripeness and postharvest life. The USDA Color Chart allows human graders to classify the tomatoes into six maturity stages: Green, Breakers, Turning, Pink, Light Red, and Red. Similarly, the most common practice of grading eggs in most backyard egg farms is through visual inspection of the egg shell surface. Expert human graders look for the presence of pin holes, stretch marks, strain, discoloration, and dirt on eggs. The egg is rejected for marketing if pin holes, stretch marks, strain or discoloration is present and is accepted if only dirt is present as dirt can be cleaned. The advantage of using humans for grading and classification of agricultural produce is that humans can visually ``memorize" the object's external characteristics, while at the same time can easily adapt to subtle differences in the grader's environment (eg., brought about by the effect of lighting condition of the workplace) as well as personal differences (e.g., boredom and eye stress) without losing accuracy. Being able to memorize a complex combination of object characteristics and being adaptive at the same time are human attributes that are very difficult to model for automation purposes.

Several procedures have already been devised by researchers and developers to automate the sorting process in natural and agricultural food products. Recenty, the most widely-used method is by the use of laser-based spectroscopy to assess the maturity of fruits~\citep{Gunasekaran90, Moya05, Forbus06a, Forbus06b, Clement08}. In this method, the intensity of the delayed light emission (DLE)~\citep{Strehler51} was correlated to chlorophyll and carotenoid pigment contents of the fruit, which are indicative of maturity stage~\citep{Forbus06a, Forbus06b}. In the case of grading eggs, on the other hand, the currently available state-of-the-art is via a color computer vision system with an expert system (ES) that uses the knowledge-base learned by an artificial neural network (ANN) to  sort the eggs~\citep{Patel98b, Mertens05}. Here, the ANN is trained to identify blood spot, crack and dirt stain defects under a controlled environment~\citep{Patel98b, Mertens05}. Using DLE for classifying fruits requires procuring expensive devices such as the laser source in laser-based spectroscopy, while the ES will only work in a controlled  environment different from the current human working environment. Locally implementing these state-of-the-art solutions would require a greater amount of capitalization for the farm owners, as well as trial-and-error-based customization for the developers.

A MVS is a system composed of robust algorithms that integrate state-of-the art software technologies and (sometimes cheap) hardware. The MVS hardware is composed mainly of a camera and a computer. The camera serves as the ``eye" of the MVS that senses and then captures the visual images of objects. The computer runs the ``brain" of the MVS which outputs the classification of the objects. The brain of the MVS is an ANN that can be trained to mimic the decision-making capabilities of humans. An ANN is an abstract model of the human brain that is proven to be robust in dealing ambiguous and noisy classification problems.  Like the human brain, it is capable of interpolating and extrapolating large amount of data and can simultaneously explore many hypotheses using massive parallelism~\citep{DeGrano07}.

In this effort, we developed a cheap, off-the-shelf MVS by integrating a locally-available web-camera and a computer for the purpose of providing local farmers an alternative to manual grading. We designed our MVS such that they can be integrated with other online farm processing tasks, such as a sorting machine, a packing machine or a transporting machine via a conveyor belt. They can work round-the-clock and can take dimensional measurements more accurately and consistently than humans. Further, they can give an objective measure of color and morphology which a human grader could only do subjectively. The use of MVS can be hygienic to the food products because no human physical contact with the produce is involved and the possibility of damage during grading is lessen. Our main purpose in this paper, however, is to find the optimal combination of ANNs that can collectively act as the brain of the MVS and at the same time be able to mimic the grading capabilities of the human graders. 

This papers presents the development and evaluation of ANNs as classifiers of two MVS for classifying tomatoes and grading eggs, respectively. We aim to present the answer to the following question: {\it Can an ANN be developed and optimized to act as a classifier for a cheap MVS without manipulating the human environment and on the average perform as good as or better than the human graders'?} Non-manipulation of the human environment means that the system will not require changing the current working condition, such as installing more lighting devices to control light. This paper also presents: 
\begin{enumerate}
\item A methodology for measuring the performance of the human graders over an 8--hour work shift; 
\item A methodology for manipulating digital color images for inputs to the ANN; and 
\item A methodology for optimizing the ANN structure. 
\end{enumerate}
Our evaluation results show that 97.00\% of the 1,200 tomato images under the evaluation set were correctly classified by the MVS with the 3.00\% misclassified images within one maturity stage difference. Similarly, 75.89\% of the 113 egg images were correctly classified by the  MVS where among the misclassified eggs, only 6.25\% were false positives.

\section{Review of Literature}
\subsection{Current State-of-the-art in Tomato Maturity Classification}
The MVS has already been used by several researchers for automatic sorting of tomatoes~\citep{Watada76, Slaughter96, Hong98}. In general, their respective systems use spectroscopy techniques to analyze the internal characteristics of the fruit such as acidity, soluble solids content, pH, maturity index and dry matter content. The common physical principle used in these systems is that the DLE of tomato changes as it matures so that the changes in visible spectrum and the maturity stage can be correlated. These systems, however, use expensive hardware components that are financially limiting for local adaptation, specifically the small to medium backyard farm types. 

\subsection{Current State-of-the-art in Egg Grading}
The MVS has already been used by several researchers for the assessment of different qualities of eggs. \citet{Elster91} developed a program that isolates the Grey-scale images of eggs from the background noise and successfully finds cracks from the image 96\% of the time. They extended their work to detect cracks at any point on the surface of rotating eggs but the identification process now depends on the egg size, a rotating mechanism, and the program requires some calibration~\citep{Goodrum92}. Moreover, the egg images in their setup were taken under a specific lighting condition, and specific camera position and distance. \citet{Patel98a} utilized the image-processing routines of~\citet{Elster91} to capture Grey-scale images of cracked and ``sound" eggs as inputs to ANN for the detection of cracks in eggs. Their work had an accuracy of 90\%. They later extended their work to detect blood spots and dirt stains with an accuracy of 86.5\% and 80\%, respectively\citep{Patel98b}. These systems, in general, depend on an extra mechanism along the grading line (e.g., a rotating mechanism), the camera resolution, the lighting condition, the number of variables measured on each egg, the algorithm used for extracting features, the shell color, and the size and type of egg defect. Further, their ANNs were trained to grade eggs based only on one or two characteristics and not all characteristics that the human graders use. Their ANNs were also trained using images that were taken under a controlled environment (i.e., controlled lighting, controlled camera angle and distance). 

\section{Scientific Framework}
\subsection{Artificial Neural Networks}
An artificial neural network (ANN) is a nature-inspired computational paradigm whose structure and processes model how biological nervous systems process information. The ANN is composed of a large number of processing nodes, aptly called neurons, that are highly interconnected by weighted links, aptly called synapses. The structure of the ANN resembles that of the human brain and makes it able to learn by example, through a process similar to how human brains learn: adjustment of the synaptic connections between two neurons~\citep{Smith93, Haykin99}. Because of its ability to handle complex, noisy, multi-modal, multi-dimensional, discontinuous, and mixed-nominal data, ANN has been used by many researchers for learning, adapting, generalizing, clustering or organizing patterns to solve real-world problems (see for example the works of~\citet{Krose96},~\citet{Jain00} and~\citet{Rabunal06}).

ANN solves problems differently than the usual algorithmic approaches common in mathematics, engineering and management sciences. Algorithmic approaches follow a set of instructions in order to solve a problem. The difficulty of using algorithmic approaches is that only the known steps have to be followed in order to solve the problem. Thus, only understood problems with known unambiguous solution steps can be solved by algorithmic means. ANN, on the other hand, uses unconventional approaches to finding solutions by itself, such as deriving meaning from complicated or imprecise data, and  extracting patterns and detecting trends that are too complex for humans to be noticed~\citep{Smith93, Haykin99}. Because of these capabilities, ANN can be used to classify objects based on the complex combination of object characteristics that is otherwise difficult to model using algortihmic means.

\subsection{Artificial Chemistry Optimization}
Artificial Chemistry (AChem) is a computational paradigm for search, optimization, and machine learning. In this paradigm, the artificial molecules represent machine or data and the interactions among these molecules are driven by an algorithm. The duality of the molecule to represent either a machine (or operator) or data (or operand) enables a molecule to process other molecules or be processed. This dualism property of molecules enables one to implicitly define a constructive computational procedure using the dynamics of chemical reaction as a metaphor to solve complex optimization problems~\citep{Pabico03, Pabico04, Pabico06a, Pabico06b}. The capabilities of AChem to simultaneously find solutions to different problems makes it a potent solution to finding the best ANN structure for the classification problem.

\section{Objectives}
The general objectives of this research is to develop and evaluate an ANN as a classifier of a MVS for classifying the maturity of tomatoes and for grading eggs. The specific objectives are:
\begin{enumerate}
\item To construct the respective MVS using cheap, off-the-shelf hardware;
\item To develop a methodology for processing images as inputs to ANN;
\item To optimize the structure of the ANN; and
\item To evaluate the accuracy of the optimized ANN in classifying the maturity of tomato and in grading eggs.
\end{enumerate}

\section{Methodology}
\subsection{Materials}
In order to construct a computer-based vision system for classifying objects using an algorithm, we integrated a cheap web-camera with a simple desktop computer running an open-source multiprogramming operating system (OS). The MVS for tomato maturity classification used a Logitech QuickCam Communicate STX web-camera and a Pentium IV desktop PC running the Gnu/Linux OS. Images of 6,000 fresh tomatoes harvested in Summer  2003 from the Netafim Farm in Tagaytay City were taken using the web-camera. These images equally represent the six maturity stages: 1,000 images each for  Green, Breakers, Turning, Pink, Light Red, and Red maturity stages, respectively.

\begin{figure}[htb]
\centering\epsfig{file=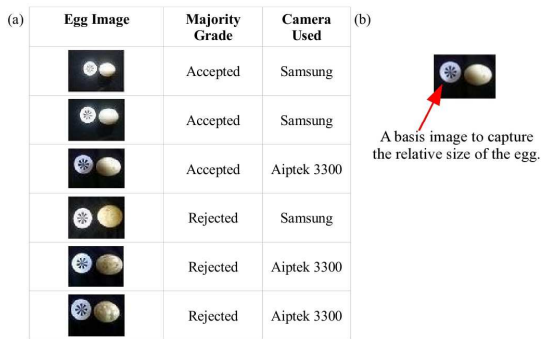, width=2.8in}
\caption{(a) Sample egg images with the corresponding grade and the brand of camera used; and (b) Each image was taken beside a basis image to capture the relative size of the egg.}\label{fig:egg-sample}
\end{figure}

The MVS for egg grading used either the Aiptek 3300 or Samsung web-cameras connected to a desktop PC with AMD Athlon 64-bit processor running the Gnu/Linux OS. Images of 750 eggs harvested in October 2005 from a backyard poultry farm in Sto. Cristo, Sariaya, Quezon were collected using the system. Of the 750 images, 50\% were positive examples (i.e., 375 were graded as acceptable) while the remaining 50\% were negative examples (i.e., graded as rejects by the farm's in-house egg graders). Sample egg images with their corresponding image acquisition data are shown in Figure~\ref{fig:egg-sample}. For scaling purposes, the image beside the egg provides a baseline for measuring the size of the egg.

\subsection{Measuring the Performance of Human Graders}
To compare the performance of the MVS with that of the human graders, we measured the grading accuracies of the human graders over an 8--hour day shift. In this measurement, we assumed that the human graders were performing at 100\% during the first hour of the shift (at 8:00 A.M.). We have taken 120 tomato fruits classified by five independent graders during the first hour of the shift (equally representing the 6 maturity stages) as the benchmark tomatoes. We carefully tagged each fruit and their respective maturity classification recorded. The tags, however, were not visible to the graders. Also, the human graders did not know that their respective performance was being measured (i.e., to avoid artificial increase in their grading performances due to the {\it Observer's Paradox} or {\it Hawthorne's Effect}~\citep{Franke78,McCarney07}). During the second hour, and every hour thereafter, we mixed the same set of tomatoes with the tomatoes to be graded during that hour, and then we allowed them to be re-graded by the five graders. In this set, we recorded the grade of each tomato on each hour. We conducted a similar measurement on the performance of the three human egg graders on 100 eggs, of which 50 were graded as accepted and 50 as rejects during the first hour. 

\subsection{Image Acquisition}
\begin{figure}[htb]
\centering\epsfig{file=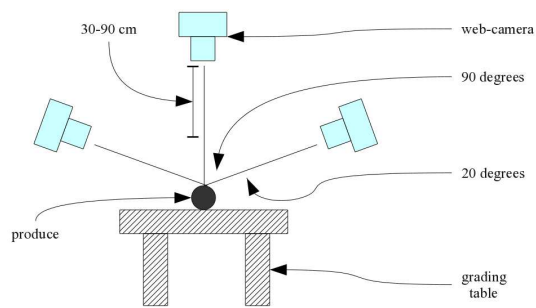, width=2.8in}
\caption{The camera set-up for acquiring the images of tomatoes and eggs.}\label{fig:angle}
\end{figure}

 Using the two respective MVS web-cameras, we acquired 6,000 images of tomato fruits and 750 images of eggs. To capture the varying lighting condition of the human graders, we took the images under the same lighting condition as the human graders over the 8--hour shift. Thus, the collected images have varying lighting shades. We acquired the images of the produce  within the range 30 cm to 90 cm from the camera lens, with the angle of acquisition ranges from 20 degrees to 90 degrees with respect to the horizontal grading table (Figure~\ref{fig:angle}). We saved the images in JPEG format, together with a file that identifies the respective classification of the products.

\subsection{Processing Images for Inputs to ANN}
The following procedure were performed on each tomato and egg image:
\begin{enumerate}
\item Using a simple edge-detection algorithm~\citep{Canny86, Deriche87}, we automatically identified and extracted the image of tomatoes and eggs by removing their respective backgrounds.
\item We extracted the red, green, and blue (RGB) spectral patterns of the background-less image to make sure that only the RGB patterns of the produce were recorded.
\item We normalized the RGB patterns by scaling each RGB value with the number of pixels of the product image. Thus, the RGB patterns have values that range from 0~to~1.
\item We then stored the normalized RGB patterns in a file together with the image's classification.
\end{enumerate}

\subsection{ANN Optimization and Training}
Finding the optimal ANN structure for a given classification problem is a problem that faces a combinatorial explosion. This problem becomes intractable for very large input instances. For our classification problem, the number of input neurons for both tomato maturity classification and egg grading problems is $768$ ($3\times 256$), representing the frequency of each of the $256$ normalized red, green and blue values. The number of output neurons for tomato maturity is six, representing the six maturity stages, while that for the egg grading is only one, representing the binary output of accept and reject  grades. Our ANN structuring problem is a 4--factor optimal combination: i.e., To find the optimal number of layers in the hidden layer, the optimal number of neurons on each hidden layer, whether or not the ANN will be structured as a jump-connection nets~\citep{Smith93, Haykin99}, and the type of activation function that each neuron will use. 

We setup an AChem system to simultaneously find the (near-)optimal ANN structure for tomato maturity classification and egg grading. We allowed the AChem system to run for either 10,000 simulation cycles or when 80\% of the molecules already encode the same ANN structure. Each AChem molecule encodes the number of hidden layers, the number of neurons on each hidden layer, a binary digit that flags whether the structure will use a jump-connection or not, the activation function type, the learning rate, and the momentum value. We devised specialized reaction rules such that the collision of the two molecules, as well as the collision of a molecule with the artificial reaction tank, will create more molecules. We also devised a reactor algorithm to filter out molecules that encode better solutions. The details of our AChem system have been presented elsewhere~\citep{DeGrano07, Pabico07}.

We evaluated each encoded ANN structure on AChem's molecule by running a feed-forward, back-propagation learning algorithm~\citep{Smith93} over the training data using the accompanying encoded learning rate and momentum. For the ANN for tomato maturity classification, we used 4,200 images as the training set, with each maturity stage equally represented (700 images per maturity stage). In our ANN for egg grading, we used 526 images as the training set, containing 263 images with accepted grade and 263 images as rejects. We extracted a test set, disjoint from the training set, containing 600 images for tomato classification (each maturity stage equally represented) and 112 images for egg grading (each grade equally represented). We stopped the ANN training when the network error over the test set did not improve after 100 epochs. We considered the ANN as trained when the network error over the test set was at the minimum. Similarly, we extracted a validation set, disjoint from both the training and test sets, containing 1,200 images for tomato classification (each maturity stage equally contains 200 images) and 112 images for egg grading (56 images representing the accepted grade and 56 for the rejects). We ran the trained ANN over the validation set and its classification rate recorded. The recorded classification rate becomes the molecule's ``molecular weight" (i.e., fitness value). 

\section{Results and Discussion}
\subsection{Performance of Human Graders}
\begin{figure}[hbt]
\centering\epsfig{file=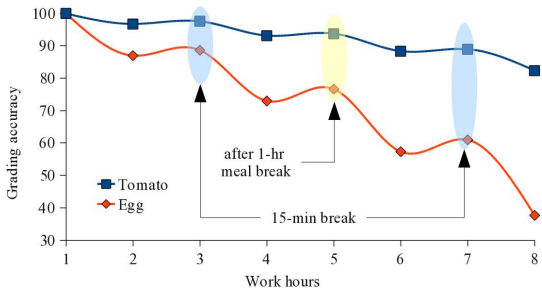, width=2.8in}
\caption{Average performance of 5 tomato graders and 3 egg graders over an 8--hour work shift.}\label{fig:graders-performance}
\end{figure}

Figure~\ref{fig:graders-performance} shows the average accuracy of five tomato graders and three egg graders over an 8--hour work shift. The respective grading performance of the tomato and the egg graders show a similar decreasing pattern over time, with sudden peaks after each of the three breaks (two 15-minute breaks and one 1-hour meal break). The daily average grading accuracy of the human graders was 92.65\% for tomato graders and 72.67\% for egg graders.

We atrribute the observed decreasing performance pattern of the human graders to the physical limitations of humans such as tiredness, eye stress and boredom due to the repetitive nature of the work. The performance increase after each and every break maybe attributed to the graders being able to take a rest, thus their eyes were refreshed while their brains were refocused to other functions. However, the break was not enough to bring the humans' performance accuracy back to its 100\% level. The average grading accuracy of the egg graders decreases more rapidly over the 8--hour shift than the tomato graders. The reason for this is that the color variability in eggs is less than that of the tomatoes which could have rapidly increased the egg graders' boredom and eye stress.

\subsection{Optimized ANN for Tomato Maturity Classification and Its Performance}
Our AChem optimization routine found the optimal ANN structure for tomato maturity classification as a 4--layer feed-forward network with 768 neurons on each of the two hidden layers. The neurons used the sigmoid activation function. The input layer accepts the normalized RGB spectral patterns while the 6-neuron output layer outputs the maturity stage. The images in the validation set were fed forward into the network and the corresponding classification on each image was recorded. The ANN classification on each image was compared to the listed classification made by the 5 human graders. Of the 1,200 images, the ANN classified 1,164 images with the same classification as that of the human graders for a 97.00\% accuracy. Of the  36 misclassified images, the difference is within only one maturity stage. Some of the tomatoes that were classified by humans as in the Breakers stage were classified by the ANN as either in the Green or the Turning stages, while those classified by humans as in the Light Red stage were classified by the ANN as either in  the Pink or the Red stages. Based on the ANN's performance of 97.00\% grading accuracy over the validation set and the humans' average daily grading accuracy of 92.65\%, the ANN has increased the grading accuracy by 4.35\%.

\subsection{Optimized ANN for Egg Grading and Its Performance}
\begin{figure}[hbt]
\centering\epsfig{file=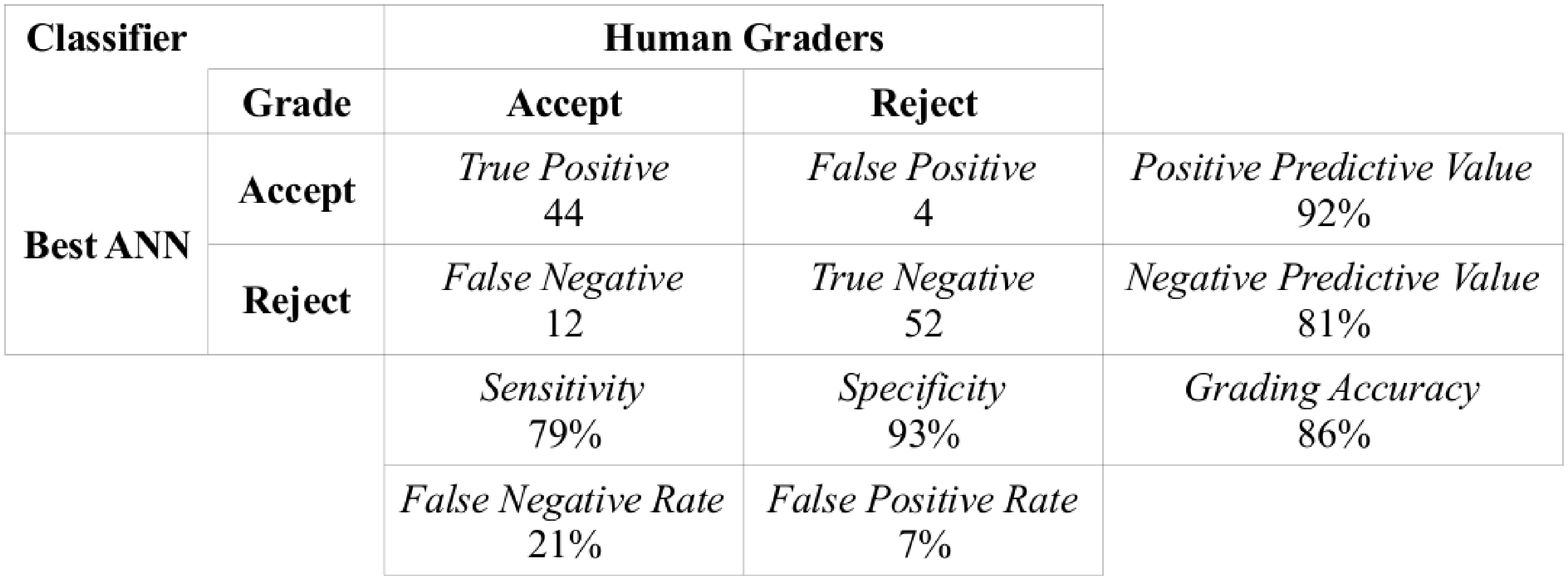, width=2.8in}
\caption{Accuracy, sensitivity, specificity, and other various performance measures of the best ANN found by AChem.}\label{fig:egg-results}
\end{figure}

The optimal ANN structure for egg grading was found by AChem optimization routine as a 3-layer feed-forward network with 768 neurons on the hidden layer. The input layer accepts the normalized RGB spectral patterns while the 1-neuron output layer corresponds to the ``accept" or ``reject" grade of each egg. The 112 egg images in the validation set were fed forward into the ANN and the corresponding grade on each image was recorded. The ANN grade on each egg image was compared to the listed grade made by the 3 human graders. Of the 112 images, ANN gave 96 images the same grade as humans for an accuracy of 86\% (Figure~\ref{fig:egg-results}). Of the 16 wrongly graded eggs, 4 images were graded as accepted but were actually rejects for a false positive rate of 7\%. The ANN's false negative rate is 21\%, grading 12 accepted eggs as rejects. The best ANN grades accepted eggs as accepted 79\% of the time (sensitivity), and rejects rejected eggs 93\% of the time (specificity). The ANN has a precision rate of 92\% (positive predictive value) and a negative predictive value of 81\%. Based on the 86\% grading performance of the ANN over the validation set and the 73\% average daily grading accuracy of the human graders, the ANN has increased the egg grading accuracy by 13\%. For an average egg farm that produces 10,000 eggs daily, the increase in grading accuracy translates to 1,300 correctly graded eggs. At the current year's price of 4 Pesos for a medium-sized egg, the increase in grading accuracy financially translates into an additional 5,200 Pesos daily revenue for the farm. This additional amount is more than enough payment for the daily minimum-wage of the three graders.

\section{Conclusion and Extensions}
In this research project, two MVSs from cheap, off-the-shelf hardware were setup and integrated with open-source software systems. The MVSs were fitted with ANNs to act as the brain for classifying the maturity of tomatoes and for grading eggs, respectively. The camera system captured images of tomatoes and eggs within the lighting conditions of human graders. To further simulate the human grading conditions, the images were captured within 30 cm to 90 cm of the camera lens and within the range of 20 through 90 degrees angle with respect to the grading table. Each of the image were stored in a JPEG file format and then subjected into image processing for input to the ANN. The image's background were removed using a simple edge-detection algorithm and the RGB spectral patterns of the background-less image were extracted. The RGB spectral patterns were normalized and then fed into the input layer of the ANN. The optimal ANN structure were found using the AChem optimization routine. The ANN for tomato maturity classification has a grading accuracy of 97\%, increasing the grading accuracy of humans by 4.35\%. The ANN for egg grading has a grading accuracy of 86\% which improved the grading accuracy of humans by 13\%. With these results, it is therefore concluded that the ANNs developed in this research is a potent classifier for MVSs for tomato maturity classification and egg grading, respectively. The systems we developed and evaluated in this research project may be used to augment, if not to replace, human graders.

The following efforts are already underway as extensions to this research endeavor:
\begin{enumerate}
\item {\bf Time and motion study to measure and compare the grading rate of humans and the MVS}. In this current effort, we did not measure the grading rate of both humans and the MVS because the MVS was not fitted with conveyor belts that will automatically feed the produce in front of the web camera. We hypothesize that the MVS grading rate will depend mainly on the speed of the conveyor belt, the speed of the camera's image acquisition, the speed of the image pre-processing procedure, and the speed of the ANN to return the grade. 
\item {\bf Utilization of newer, faster and higher resolution web cameras}. We hypothesize that newer model web-cameras of higher resolutions and faster image acquisition will improve the performance of the MVS. These camera models have already been made available in the market with prices lower than the web cameras used in this study.
\item {\bf Minimizing the image features used as input to the ANN}. Minimizing the number of nodes in the input layer of the ANN may increase the MVS grading efficiency. We hypothesize that the use of gray-scaled versions of the color images will not affect the grading accuracy of the MVS, but in turn will decrease the required number of input nodes by 33\% from 768 normalized RGB shades down to 256 normalized shades of gray.
\item {\bf Utilization and comparison of various techniques for creating a committee of ANNs}. With the advent of new and novel advances in ANN research, we are currently looking at the possibility of using a committee of ANNs~\citep{Tresp01, Dietterich02} to increase the grading accuracy further. Various committee machine techniques have already been developed and tested by different researchers such as averaging, bagging, boosting, and adaptive boosting. Comparing the performance of the MVS when using these techniques to the performance of our current MVS must be performed to further increase the respective systems' grading accuracies.
\item {\bf Development, implementation, evaluation and optimization of MVS as a digital sensor for other agricultural products}.  The development of the respective MVS for the following products are underway: mangoes, abaca, and feed quality.
\end{enumerate}



\section{Acknowledgments}
We thank the following:
\begin{enumerate}
\item Lim's Poultry Farm, Sto. Cristo, Sariaya, Quezon where the eggs came from; and
\item Netafim Farm, Tagaytay City where the tomatoes came from;
\end{enumerate}

\bibliographystyle{plainnat}
\bibliography{paper1}


\end{document}